# BLACK-BOX MODELLING OF HVAC SYSTEM: IMPROVING THE PERFORMANCES OF NEURAL NETWORKS

Eric FOCK[(a)], Thierry A. MARA[(a)], Philippe LAURET[(a)] and Harry BOYER[(b)]
Laboratoire de Génie Industriel – Equipe Génie Civil et Thermique de l'Habitat[†]
[(a)] Université de La Réunion – 15, av. René Cassin – BP 7151 – 97715 Saint Denis Cedex 9
[(b)] Institut Universitaire de Technologie – 40, av. de Soweto - 97410 Saint Pierre
Ile de La Réunion - FRANCE

## ABSTRACT

This paper deals with neural networks modelling of HVAC systems.

In order to increase the neural networks performances, a method based on sensitivity analysis is applied. The same technique is also used to compute the relevance of each input.

To avoid the prediction errors in dry coil conditions, a metamodel for each capacity is derived from the neural networks. The regression coefficients of the polynomial forms are identified through the use of spectral analysis.

These methods based on sensitivity and spectral analysis lead to an optimized neural network model, as regard to its architecture and predictions.

## NOMENCLATURE

**Symbols**

| | |
|---|---|
| Nc | Number of hidden neurons |
| Ne | Number of input variables |
| $Q$ | Cooling load (e.g. total, sensible, latent, absorbed) |
| Tanh | Hyperbolic tangent |
| T | Temperature |
| $\tau$ | Constant time at the system starting up |

**Indices and Exponents**

| | |
|---|---|
| i | Input layer |
| j | Hidden layer |
| k | Output layer |
| b | Bias |
| $w_{jk}$ | Weights between the unit in j and k layers. |
| odb | Outdoor Dry Bulb |
| edb | Entering Dry Bulb |
| ewb | Entering Wet Bulb |
| ss | Steady state |

## INTRODUCTION

The performance of HVAC system can be modelled using manufacturer design data presented as derived performance maps. It is up to the modeller to predict the total, sensible and absorbed steady state capacities. These parameters are usually given for several indoor and outdoor conditions ($T_{odb}$, $T_{edb}$ and $T_{ewb}$).

The dynamic behaviour can then be derived from the steady state using a first order model.

The steady state performances can accurately be assessed using black box modelling such as neural networks. Nowadays, neural networks are widely used to model nonlinear relationship between the data. However, even a black-box model needs few parameters to achieve its goal; the performance of its prediction depends on them. A methodology based on sensitivity analysis is employed to improve the performance of the HVAC neural model. Improvements are obtained by determinating the optimal size of the hidden layer.

It is also necessary to solve the dry coil condition problem, when the sensible cooling load predicted can be higher than the total one. We present a way to overcome this problem using spectral analysis. This method allows us to obtain a metamodel of the neural network. The metamodel used here is a polynomial form of the neural model, thus allowing the dry coil condition balanced by the equation $P_{tot} = P_{sens}$ solved in a simpler way, then providing the 'real' value of $T_{ewb}$ and the adequate $P_{tot}$ and $P_{sens}$.

## ARTIFICIAL NEURAL NETWORKS

### Background

Artificial neural networks are widely being used in several fields ranging from system modelling (Kalogirou, 2000) to control (Norgaard, 1996). It has been shown that the neural networks are universal function approximators. They can approach in any function with any accuracy provided that a sufficient number of hidden neurons are used (Hornik *et al.*, 1989). A common type of neural network, the multilayer perceptron, is used in this paper (see *Figure 1*),.

---

[†] Email: (eric.fock, thierry.mara, philippe.lauret, harry.boyer) @univ-reunion.fr

The neural network output is described by

$$y = \sum_{j=1}^{Nh} w'_{jk} f_j (\sum_{i=1}^{Ne} w_{ij} x_i + b_i) + b'_j \quad (1)$$

where y is the model output and the w, w', b, b' are the parameters.

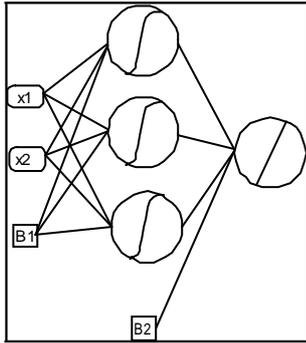

*Figure 1:Multilayer perceptron*

A classical neural network with two inputs, two biases, three units in the hidden layer and one output. The activation function in hidden units is *Tanh*.

Nevertheless, several parameters must be fixed in order to achieve the goal: number of hidden units, weight decay or another learning and pruning parameter (Reed, 1993). There is no rule to ensure the optimal amount of hidden units. A wide network, with too many hidden units, will be able to fit high nonlinearities; unfortunately, it will also perfectly learn the noise, decreasing the prediction performance, its generalization ability. A trade off must be found. The network should have enought hidden neurons to properly model the behavior of the phenomenon and to reject the noise.

**Overview of the training process**

The learning process can be summarized as follow:
1. Select the input variables,
2. Select the size of the hidden layer,
3. Select the learning parameters, such as learning error and weight decay, number of training cycle (epochs),
4. Use the error back-propagation to modify the weights,
5. Repeat the sequence over all patterns of the learning database and all epochs,
6. Stop the procedure when the accuracy is obtained or the number of epochs has been reached.

Training neural networks usually needs two databases: one for the learning step and the other for the generalization step. The latter measures the prediction performance of the trained neural network. The classical method is the trial-error stage: several networks are tested with different number of hidden units and/or learning parameters and the better, in sense of the prediction performance, is selected. One can improve the training stage using cross-validation or pruning techniques. The cross-validation is not a simple one because it needs three databases: one for the learning, another one for the generalization and the last one for the validation. It is used to stop the learning when the prediction error on the validation database begins to increase just after a decreasing period. This behaviour is schematically represented in *Figure 2*.

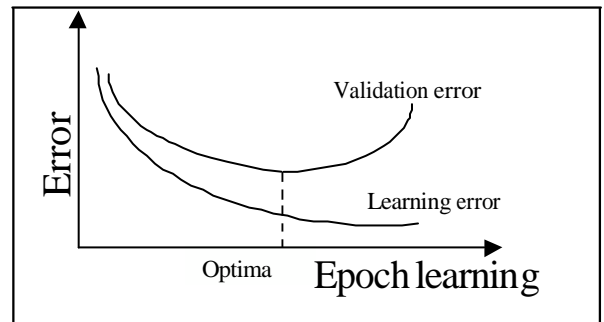

*Figure 2: Schematic view of the cross-validation*

At the optimal point, the learning should stop as the error of the prediction on the validation set begins to increase although the learning error is still decreasing.

The main problem here is of having enough data for all the three databases.

The pruning techniques aims to delete the nodes which are non-relevant, but the main problem is to fix the weight-decay parameter.

A novel pruning algorithm, based on sensitivity analysis is applied here.

**Sensitivity Analysis**

The main advantage of this method is that it is model independent and thereby applied to any multilayer perceptron network. The sensitivity analysis used here is EFAST (Extended Fourier Amplitude Sensitivity Test) (Saltelli *et al.*, 1999).

The pruning method is performed after the training stage. Hence, there is no need to modify or employ a special training algorithm

By achieving all the computations, the total sensitivity coefficients for each hidden units is available. It is a measure of the relevance of the neuron with respect to the output model.

Applying this methodology to neural networks will reduce the amount of hidden units and test the saliency of the inputs.

## HVAC MODELLING

These methodologies are applied to improve the performance of a black-box model and to solve the problem that occurs in dry coil conditions. The aim of this improvement is to reduce the model complexity

(the number of weights), i.e. to find an optimised neural network, for a given accuracy.

There are two main model classes for modelling HVAC systems. The first one describes all the components of the heat exchanger (evaporator, condenser, fluid characteristics, etc…). The second one predicts the cooling loads accordingly the environmental variables (indoor and outdoor).

The split-system considered here is modelled by using the second way. The dynamic behaviour is derived from the steady-state one and described by a first order model for both total and sensible capacities:

$$Q_{tot,cyc} = Q_{tot,ss}(1-e^{-t/\tau})$$
$$Q_{sens,cyc} = Q_{sens,ss}(1-e^{-t/\tau}) \quad (2)$$
$$Q_{abs,cyc} = Q_{abs,ss}$$

It is assumed that the absorbed capacity reaches its steady state value and the performance maps (in steady state) are functions of $T_{odb}$, $T_{edb}$ and $T_{ewb}$. These variables represent the outside temperature, the dry inside air temperature and the wet inside air temperature respectively. $T_{ewb}$ is about air state, accounting for air humidity and air temperature. The key modelling stands for the prediction of the steady state capacities. Mathematical regression can be a solution. For example, non-dimensional regressions of a wide range of systems have been made by (Garde *et al.*, 2002).

As some manufacturer's data are available, artificial neural networks are used to perform the prediction of the total, sensible and absorbed cooling loads (Fock *et al.*, 2000)

A wide neural network (with a large amount of neurons in the hidden layer) is considered, to take into account of most of the nonlinearities that involve in this phenomenon. Then, the architecture (i.e., the number of hidden units) of the network is optimised and the relevance of all the inputs is also verified.

**Improving the neural architecture**

Neural networks are the black-box method, selected to model the system capacities as performance maps. As seen previously, the modeller's attention is focused on the optimisation of the architecture. The sensitivity-based method is applied to remove the irrelevant hidden neurons. The learning parameters (number of epoch, weight-decay or others pruning parameters) remain unchanged during all the simulations. The only parameter that changes is the number of hidden units. The database properties for the learning (App) and the generalisation (Gene) stages are resumed in *Table 1*.

*Table 1: databases characteristics*

| | Values for $T_{odb}$ |
|---|---|
| App | Several except 18.33, 29.44 and 40.56°C |
| Gene | Several including 18.33, 29.44 and 40.56°C |

| | $T_{odb}$ (°C) | $T_{edb}$ (°C) | $T_{ewb}$ (°C) |
|---|---|---|---|
| Min | 12.78 | 12.78 | 4.4 |
| Max | 46.11 | 35.00 | 35.00 |

| $T_{odb}$ (°C) | $T_{edb}$ (°C) | $T_{ewb}$ (°C) | $P_{tot}$ (kW) | $P_{sens}$ (kW) |
|---|---|---|---|---|
| 40.56 | 32.22 | 18.33 | 33.17 | 33.17 |
| 40.56 | 32.22 | 21.11 | 33.67 | 32.13 |
| 40.56 | 32.22 | 23.89 | 36.13 | 26.86 |
| 40.56 | 32.22 | 26.67 | 39.18 | 20.91 |
| 40.56 | 32.22 | 29.44 | 42.24 | 14.94 |
| 40.56 | 32.22 | 32.22 | 45.27 | 9.03 |

For example, the table above shows some values of the generalisation database.

For a given value of $T_{odb}$ (40.56°C) that has not been used during the learning stage, and for a given value of $T_{edb}$ (32.22°C), there are several values of $P_{tot}$ and $P_{sens}$, according to $T_{ewb}$.

One can see that for $T_{ewb} = 18.33°C$, $P_{tot} = P_{sens}$. This means: there is no dehumidification, the split-system operates in dry coil conditions.

For the databases used during the training stage, the cited values of $T_{odb}$ (e.g. 40.56°C) means that no values for $T_{odb} = 40.56°C$ were available.

A large multilayer perceptron is set, with the following characteristics:
- 20 hidden units (Nc = 20)
- activation function $f(\bullet) = tanh(\bullet)$
- 3 inputs: $T_{odb}$, $T_{edb}$ and $T_{ewb}$
- 1 output

The neural network complexity is described by
$$(Ne+1)*Nc + (Nc+1)*1 \quad (3)$$
i.e. 101 parameters.

This skeleton of neural model is used for each of the three capacities: $P_{tot}$, $P_{sens}$ and $P_{abs}$.

A learning stage for each neural network is performed in order to fix the weights. The back-propagation algorithm (Bishop, 1995) is used with the learning database and then, the prediction performances of the model are tested on the generalisation dataset. *Figures 3a and 4a* show the histogram of the relative errors for two models.

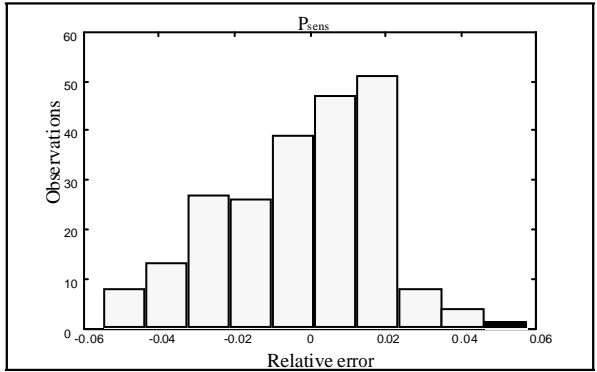

*Figure 3a: Sensible capacity prediction with Nc=20 hidden units*

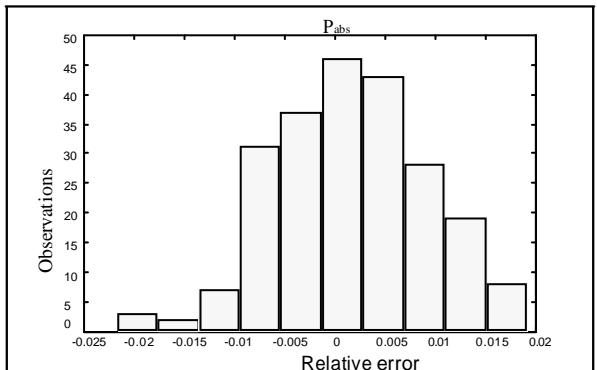

*Figure 4a: Absorbed capacity prediction with Nc=20 hidden units*

In the second step, sensitivity analysis is applied on the models in order to reduce the quantity of hidden units and thus the number of parameters, while keeping the same degree of accuracy (Lauret *et al.*, 2003).

*Figures 5-6* show the sensitivity coefficients for the 20 neurons in the hidden layer.

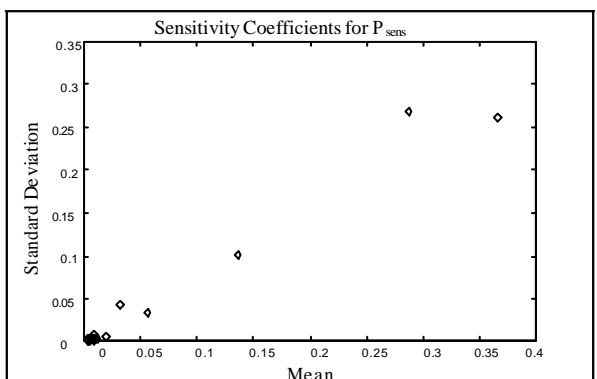

*Figure 5: Sensitivity of hidden units for sensible capacity*

The figure stands out five neurons. With a very low (near zero) mean and standard deviation, these neurons are useless.

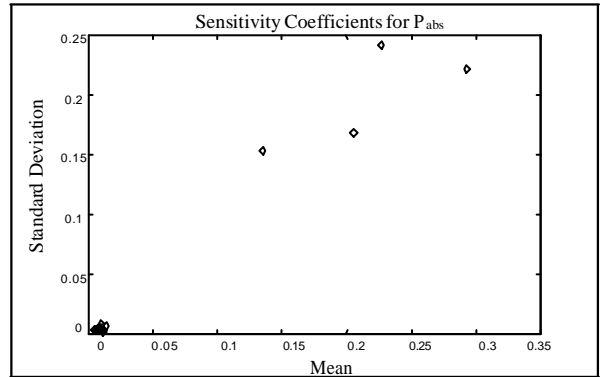

*Figure 6: Sensitivity of hidden units for absorbed capacity*

The same information is shown here. Four neurons are standing out.

Once the new architecture is selected, a new learning stage starts, using the same learning and pruning parameters.

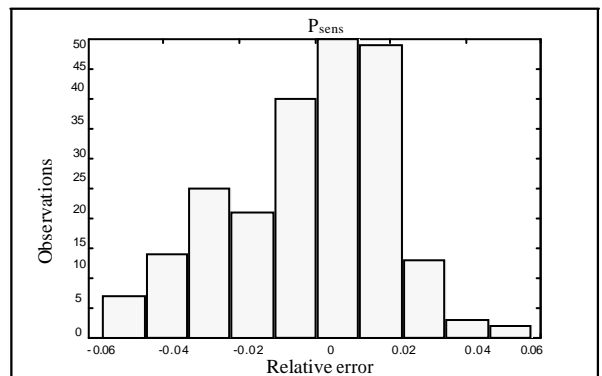

*Figure 3b: Sensible capacity prediction with Nc=5*

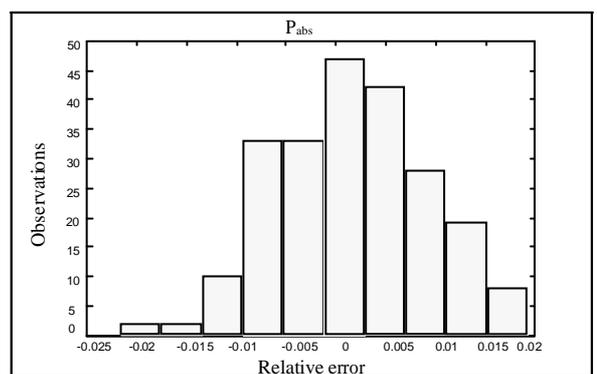

*Figure 4b: Absorbed capacity prediction with Nc=4*

As shown in *Figures 3b and 4b* and also in Table 2 below, the level of accuracy is the same, while the effective number of hidden neurons has been significantly reduced from 20 to 5 (for the total and sensible capacities).
The neural network models need only 26 parameters to achieve their goals, instead of 101.

Table 2: error in generalisation stage

|        | $P_{tot}$ | $P_{sens}$ | $P_{abs}$ |
|--------|-----------|------------|-----------|
| Nc = 20 | 5.8e-4   | 0.002      | 0.001     |
| Nc = 5  | 7.6e-4   | 0.002      | 0.002     |

The *Figure 7* shows the relative error distribution for the sensible capacity predictions in the generalisation stage.

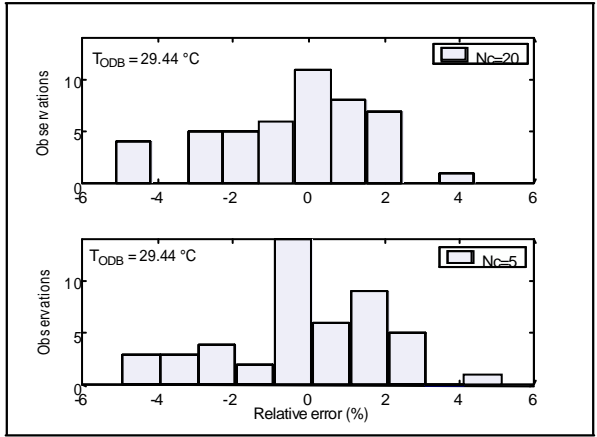

*Figure 7: Comparison of prediction performances for the sensible capacity using Nc=20 and Nc=5 hidden units*

A sensitivity analysis on the model inputs is performed to ensure that no input has been undertaken during the learning stage. All inputs are relevant, explaining more than 5% of the output variance, as shown in *Figure 8*.

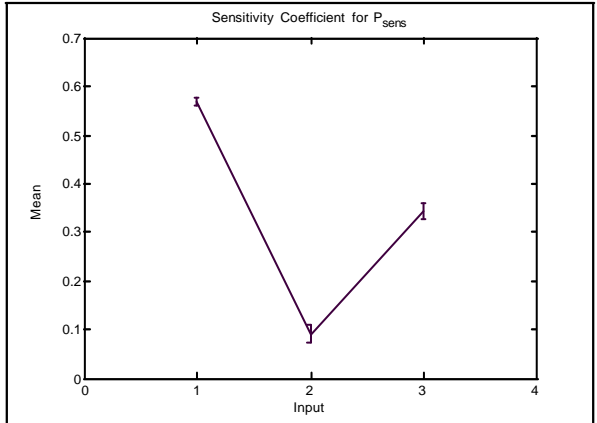

*Figure 8: Relevance of the inputs for Psens*

The neural networks models are considered to be optimised and these configurations can be used to model the steady-state cooling loads of the system.

**Overview of the black box modelling process**

The entire process is summarized as below:
1. Perform the training process using a wide number of hidden units,
2. Compute the sensitivity analysis on the neural model obtained to prune the non pertinent hidden neurons,
3. Perform the training process for the new neural network,
4. A sensitivity analysis is performed on the inputs

At the end of these stages, the neural networks architecture has been optimised regards to the problem.

**Improving prediction: dry coil conditions case**

A model of split-system using these neural networks for the prediction of the steady state cooling loads has been integrated into a thermal building software. This approach is used to model a split-system described in the HVAC BESTEST procedure (Neymark and Judkoff, 2002).
BESTEST is a comparative software procedure. It is made of several test cases, with specific configurations and specific loads schedule (Neymark *et al.*, 2002). The performances of leader softwares running all the cases are provided, allowing one to compare its own results.
During a test round, a problem occurred in dry coil conditions. In this case, the predicted sensible capacity may be higher to the total one. This situation is caused by the numerical model of the system behaviour that is not a physical one: the total cooling load is the sum of the sensible and the latent ones, both are of positive. In dry coil condition, the sensible capacity is set equal to the total one.
For example, the neural networks predictions in a dry coil condition are summarized in the Table below. One can see that the sensible cooling capacity is higher than the total one.

Table 3: performance mapping in dry coil condition

| $T_{odb}$ (°C) | $T_{edb}$ (°C) | $T_{ewb}$ (°C) | $P_{tot}$ (kW) | $P_{sens}$ (kW) |
|---|---|---|---|---|
|   |   |   | Original Data | |
|   |   |   | 33.17 | 33.17 |
| 40.56 | 32.22 | 18.33 | Neural networks predictions | |
|   |   |   | 32.84 | 33.66 |

When this case occurs, it is necessary to calculate, according to the indoor and outdoor temperatures $T_{odb}$ and $T_{edb}$, the 'real' value of $T_{ewb}$ to allow the prediction of the adequate value for $P_{abs}$ and $P_{tot}$. Remaining the mathematical form of the neural net (1), it is necessary to solve

$$P_{tot} - P_{sens} = F(T_{ewb}^{opt}) = 0 \qquad (4)$$

A simpler way to solve the above equation consists in using the spectral analysis method.

**Spectral Analysis**

By adapting the method as described by (Mara *et al.*, 2002), it is possible to determine the coefficients of a polynomial regression of the neural network. The main idea is to sample each factor so that it describes a perfect sinusoid along the entire simulation. The obtained Fourier coefficients are the coefficients of the polynomials.

It is then easier to find the root $T_{ewb}^{opt}$ for a polynomial function rather than an expression such as (1).

**Application to the dry coil conditions problem**

Polynomial forms are created from the neural networks, with the slightly same accuracy. Three polynomials are found:

$$P_{tot} = a_1 + a_2 T_{EWB} + a_3 T_{EDB} + a_4 T_{ODB} \\ + a_5 T_{EWB}^2 + a_6 T_{EDB}^2 + a_7 T_{ODB}^2 \\ + a_8 T_{EWB} T_{ODB} \quad (5)$$

$$P_{sens} = a_1 + a_2 T_{EWB} + a_3 T_{EDB} + a_4 T_{ODB} \\ + a_5 T_{EWB}^2 + a_6 T_{EDB}^2 + a_7 T_{ODB}^2 \\ + a_8 T_{EWB} T_{ODB} \quad (6)$$

$$P_{abs} = a_1 + a_2 T_{EWB} + a_3 T_{EDB} + a_4 T_{ODB} + a_5 T_{EWB}^2 \\ + a_6 T_{ODB}^2 + a_7 T_{EWB} T_{ODB} + a_8 T_{EWB}^2 T_{EDB} \quad (7)$$

Now, it is easier to solve the Eq. (4) and get the searched output $T_{ewb}^{opt}$.

The adequate values for $P_{tot}$ and $P_{abs}$ can then be computed by using the inputs $T_{odb}, T_{edb}, T_{ewb}^{opt}$.

The new values significantly affect the energy consumption, the coefficient of performance, indoor air temperature, etc.

The prediction of the behaviour model of the HVAC system has been improved for the dry coil conditions cases. The sensible capacity will never exceed the total one as well the predicted value is an adequate one, as the 'real' $T_{ewb}$, has been computed by solving (4).

The *Figures 9 and 10* illustrate the initial dry coil problem (sensible capacity is higher than the total one) and the corrected situation (sensible capacity is equal to the total predicted value and the original manufacturer data).

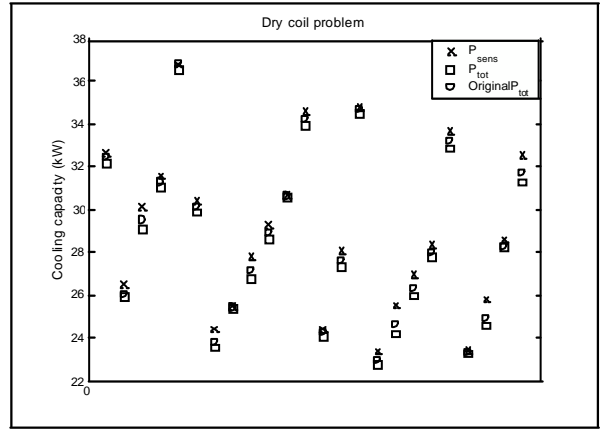

*Figure 9: Prediction in dry coil conditions*

The predicted values do not reached the manufacturer' data

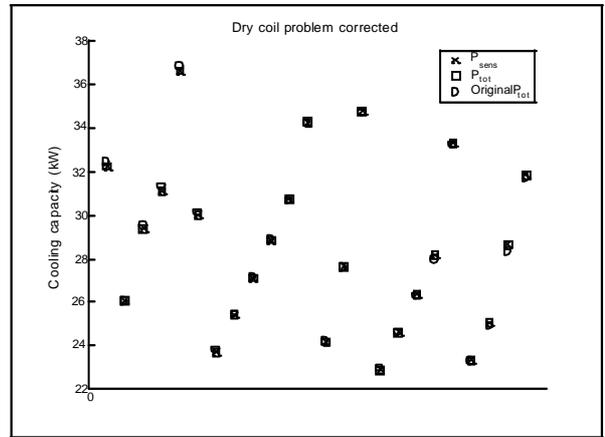

*Figure 10: Corrected predictions*

The modified $T_{ewb}^{opt}$ allows the improvements of the neural network models predictions

## CONCLUSION

A black-box model of an HVAC system has been presented here. The neural networks have been built with an optimum number of hidden neurons without any loss of accuracy by using a new pruning method. This sensitivity-based method showed that no input has been undertaken during the learning process.

In order to correct the predictions in dry coil conditions, simpler mathematical forms can be extracted from neural networks using spectral analysis. These metamodels are used to compute by an analytic way, the optimal value of the variable and then the adequate values for the outputs.

Making use of these sensitivity and spectral analysis based methods, the HVAC system model has been improved and then used during the HVAC BESTEST simulation round.

## ACKNOWLEDGMENT

Eric FOCK is the recipient of a CIFRE Fellowship funding from Electricite de France – Ile de La Reunion for which we are grateful.

## REFERENCES

Bishop, C. (1995). Neural Networks for Pattern Recognition, Oxford Press.

Fock, E., A. J.-P. Lauret, F. Garde and J.-C. Gatina (2000). "Artificial neural networks for the prediction of cooling loads of HVAC systems: a case study under tropical climate." Renewable Energy (WREC 2000) **1**: 661-664.

Garde, F., A. J.-P. Lauret, A. Bastide, T. A. Mara and F. Lucas (2002). "Development of a nondimensional model for estimating the cooling capacity and electric consumption of single speed split-systems incorporated in a building thermal simulation program." ASHRAE Transactions.

Hornik, K., M. Stinchchombe and M. White (1989). "Multilayer feedforward networks are universal approximators." Neural Networks **2**: 359-366.

Kalogirou, S. (2000). "Application of artificial neural networks for energy systems." Applied Energy **67**: 17-35.

Lauret, A. J.-P., E. Fock and T. A. Mara (2003). "A node pruning algorithm based on a Fourier amplitude sensitivity test method." submitted.

Mara, T. A., H. Boyer and F. Garde (2002). "Parametric sensitivity analysis in thermal building using a new method based on spectral analysis." Transactions of ASME - International Journal of Solar Energy Engineering **124**: 237-242.

Neymark, J. and R. Judkoff (2002). International Energy Agency Building Energy Simulation Test and Diagnostic Method for Heating, Ventilating, and Air-Conditioning Equipment Models (HVAC BESTEST), IEA - SHC Program.

Neymark, J., R. Judkoff, G. Knabe, H.-T. Le, M. Düring, A. Glass and G. Zweifel (2002). "Applying the building energy simulation test (BESTEST) diagnostic method to verification of space cooling equipment models used in whole-building energy simulation programs." Energy and Buildings **34**: 917-931.

Norgaard, M. (1996). System identification and control with neural networks. Department of Automation, Technical University of Denmark.

Reed, R. (1993). "Pruning algorithms - A survey." IEEE Transactions on Neural Networks **4**: 740-747.

Saltelli, A., S. Tarantola and K. P.-S. Chan (1999). "A quantitative model-independent method for global sensitivity analysis of model output." Technometrics **41**: 39-56.